\documentclass[runningheads]{llncs}
\usepackage[T1]{fontenc}
\usepackage{graphicx}
\usepackage{capt-of}
\usepackage{booktabs}
\usepackage{cite}
\usepackage{xcolor}
\usepackage{amsmath}
\newcommand{\red}[1]{\textcolor{red}{#1}}

\newcommand{\blue}[1]{\textcolor{blue}{#1}}

\begin{document}
\title{Fact-Checking of AI-Generated Reports} 
%
\author{Razi Mahmood\inst{1} \and Diego Machado Reyes\inst{1}\and Ge Wang\inst{1}\and Mannudeep Kalra\inst{2}\and Pingkun Yan\inst{1}}
\authorrunning{R. Mahmood et al.}
\institute{Department of Biomedical Engineering and Center for Biotechnology and Interdisciplinary Studies, Rensselaer Polytechnic Institute, Troy, NY 12180, USA
\and
Department of Radiology, Massachusetts General Hospital, Harvard Medical School, Boston MA 02114, USA}
\maketitle
%
\begin{abstract}
With advances in generative artificial intelligence (AI), it is now possible to produce realistic-looking automated reports for preliminary reads of radiology images. 
However, it is also well-known that such models often hallucinate, leading to false findings in the generated reports. In this paper, we propose a new method of fact-checking of AI-generated reports using their associated images. Specifically, the developed examiner differentiates real and fake sentences in reports by learning the association between an image and sentences describing real or potentially fake findings. To train such an examiner, we first created a new dataset of fake reports by perturbing the findings in the original ground truth radiology reports associated with images. Text encodings of real and fake sentences drawn from these reports are then paired with image encodings to learn the mapping to real/fake labels. The examiner is then demonstrated for verifying automatically generated reports.
\end{abstract}

\section{Introduction}

With the developments in radiology artificial intelligence (AI), many researchers have turned to the problem of automated reporting of imaging studies \cite{Endo2021,gale2018producing,li2019knowledge,li2018hybrid,szolovits,Pang2023,syeda-mahmood2020,xiong2019reinforced}.  This can significantly reduce the dictation workload of radiologists, leading to more consistent reports with improved accuracy and lower overall costs. While the previous work has largely used image captioning \cite{vinyals2015show,xu2015show} or image-to-text generation methods for report generation, more recent works have been using large language models (LLMs) such as GPT-4 \cite{Grewal2023,li2023artificial}. These newly emerged LLMs can generate longer and more natural sentences when prompted with good radiology-specific linguistic cues \cite{guo-2018,krause-2019}. 

However, with powerful language generation capabilities, hallucinations or false sentences are prevalent as it is difficult for those methods to identify their own errors. This has led to fact-checking methods for output generated by LLMs and large vision models (LVMs)\cite{Passi2022,nieman,midas}. Those methods detect errors either through patterns of phrases found repeatedly in text or by consulting other external textual sources for the veracity of information\cite{Passi2022,nieman,midas}. In radiology report generation, however, we have a potentially good source for fact checking, namely, the associated images, as findings reported in textual data must be verifiable through visual detection in the associated imaging.  Since most methods of report generation already examine the images in order to detect findings and generate the sentences,  bootstrapping them with an independent source of verification is needed in order to identify their own errors. 

In this paper, we propose a new imaging-driven method of fact-checking of AI-generated reports. Specifically, we develop a fact-checking examiner to differentiate between real and fake sentences in reports by learning the association between an image and sentences describing real or potentially fake findings.  To train such an examiner, we first create a new dataset of fake reports by perturbing the findings in the original ground truth radiology reports associated with images. Text encodings of real and fake sentences drawn from these reports are then paired with image encodings to learn the mapping to real or fake labels via a classifier. The utility of such an examiner is demonstrated for verifying automatically generated reports by detecting and removing fake sentences. Future generative AI approaches can use the examiner to bootstrap their report generation leading to potentially more reliable reports. 

\begin{figure*}[t]
\includegraphics[width=\textwidth]{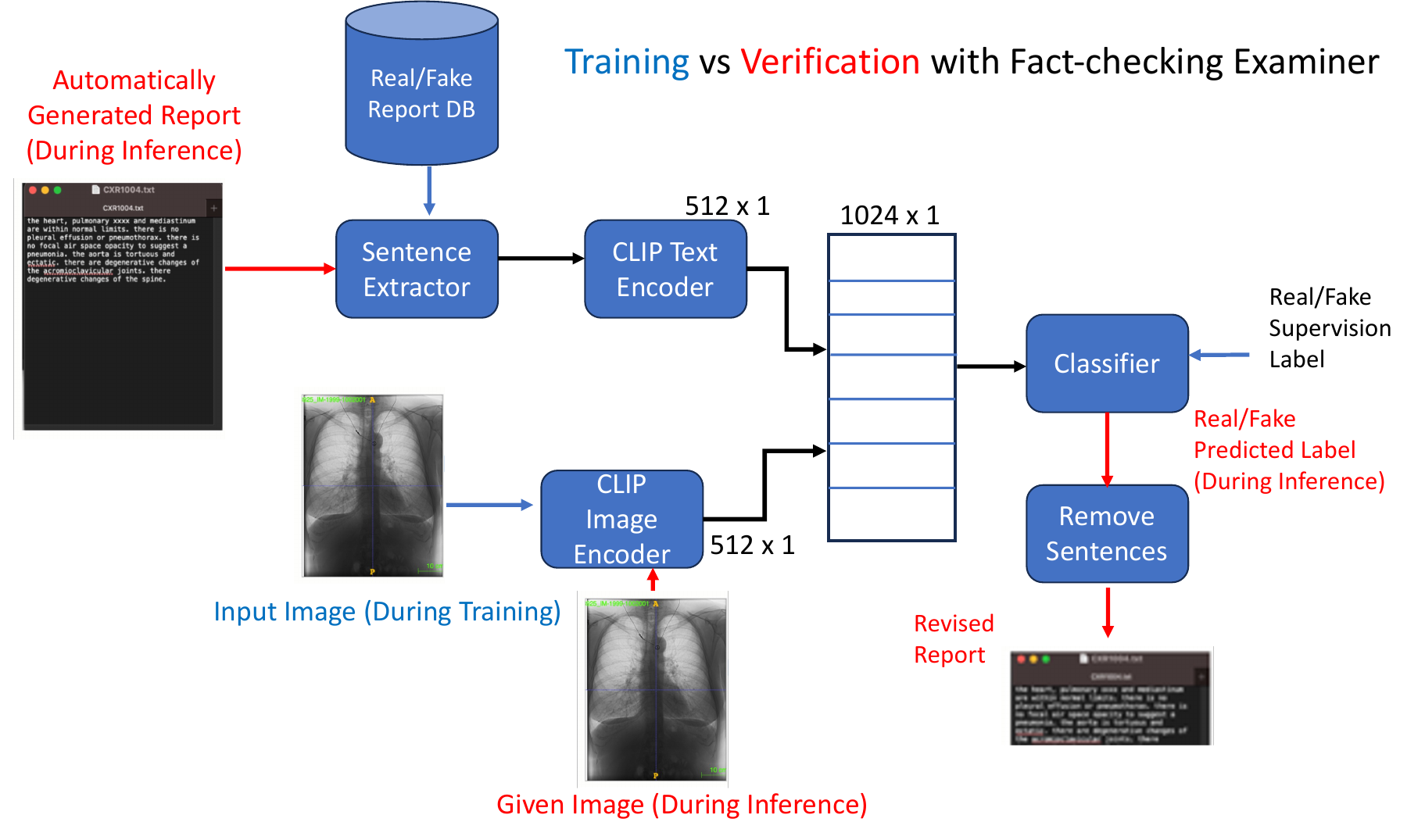}
\caption{Illustration of the training (\blue{blue}) and inference (\red{red}) phases of the image-driven fact-checking examiner. 
The operations common to both phases are in black.} \label{fig:overview}
\end{figure*}



Our overall approach to training and inference using the examiner is illustrated in Figure~\ref{fig:overview}. To create a robust examiner that is not attuned for any particular automated reporting software,
it is critical to create a dataset for training that encompasses a wide array of authentic and fabricated samples. Hence we first synthesize a dataset of real and fake reports using a carefully controlled process of perturbation of actual radiology reports associated with the images. We then pair each image with sentences from its corresponding actual report as real sentences with real label, and the perturbed sentences from fake reports as fake sentences with fake label. Both textual sentence and images are then encoded by projecting in a joint image-text embedding space using the CLIP model\cite{Radford2021}. The encoded vectors of image and the paired sentence are then concatenated to form the feature vector for classification. 
A binary classifier is then trained on this dataset to produce a discriminator for real/fake sentences associated with a given image. 

The fact-checker can be used for report verification in inference mode.  Given an automatically produced radiology report, and the  corresponding input imaging study, the examiner extracts sentences from the report, and the image-sentence pair is then subjected to the same encoding process as used in training. The combined feature vector is then given to the classifier for determination of the sentence as real or fake. A revised report is assembled by removing those sentences that are deemed fake by the classifier to produce the new report.  

\vspace{-0.1in}
\section{Generation of a synthetic report dataset}
\label{errormodeling}
\vspace{-0.05in}
The key idea in synthetic report generation is to center the perturbation operations around findings described in the finding sections of reports, as these are critical to preliminary reads of imaging studies.  
\vspace{-0.1in}
\subsection{Modeling finding-related errors in automated reports}
\label{findingtypes}
The typical errors seen in the finding sections of reports can be due to (a) addition of incorrect findings not seen in the accompanying image, (b) exchange errors, where certain findings are missed and others added, (c) reverse findings reported i.e. positive instance reported when negative instances of them are seen in image and vice versa,  (d) spurious or unnecessary findings not relevant for reporting, and finally (e) incorrect description of findings in terms of fine-grained appearance, such as extent of severity, location correctness, etc. 

\begin{figure*}[t]
\includegraphics[width=1.0\textwidth]{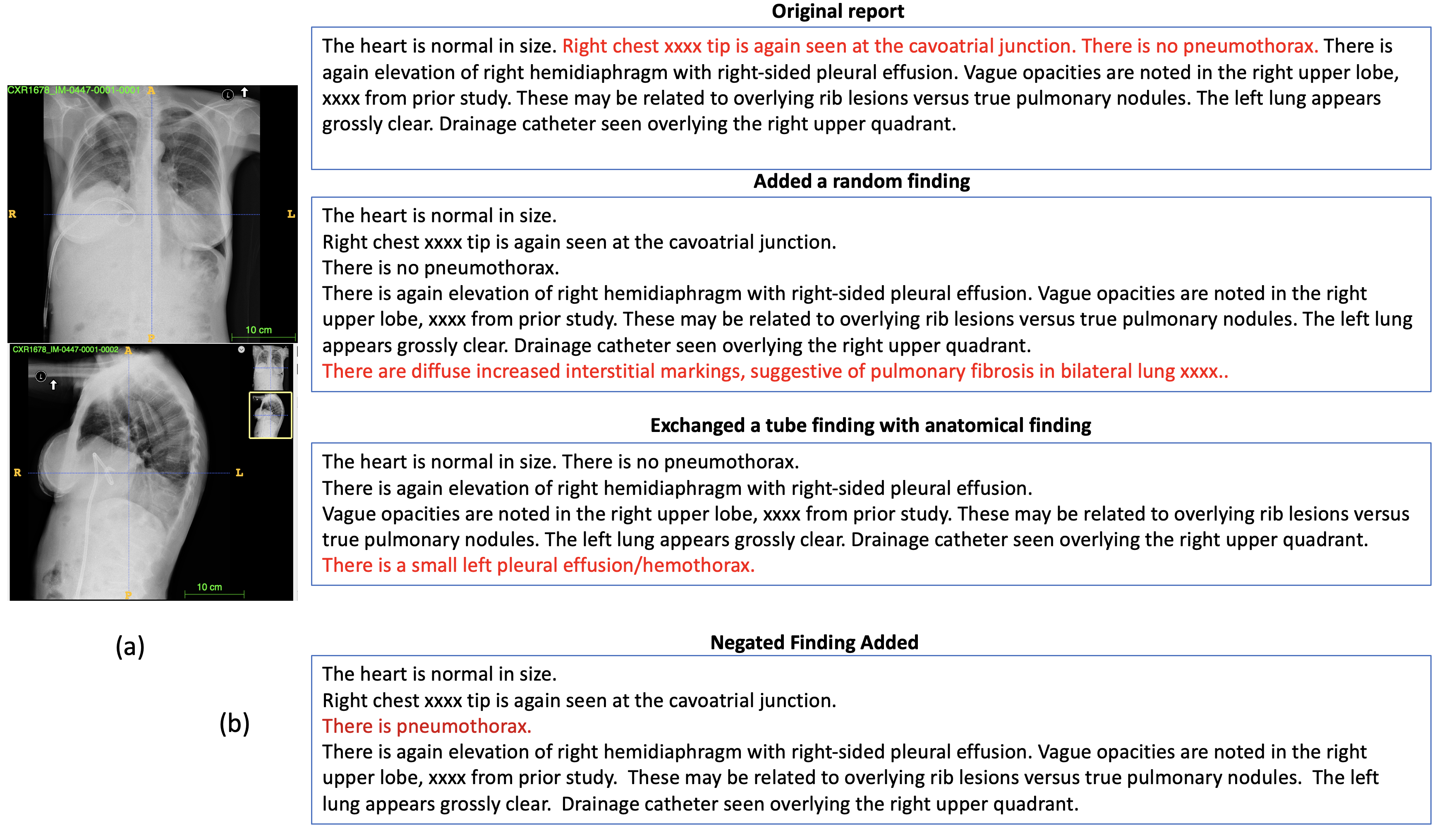}
\caption{Illustration of the fake reports drawn from actual reports. (a) Frontal and lateral views of a chest X-ray. (b) Corresponding original and fake radiology reports. The affected sentences during the synthesis operation are shown in \red{red}. } \label{fig3}
\end{figure*}

From the point of real/fake detection, we focus on the first 3 classes of errors for synthesis as they are the most common. Let $R=\{S_{i}\}$ be a ground-truthed report corresponding to an image $I$ consisting of sentences $\{S_{i}\}$ describing corresponding findings $\{F_{i}\}$.  Then we can simulate a random addition of a new finding by extending the report $R$ as $R_{a}=\{S_{i}\}\cup \{S_{a}\}$ where $S_{a}$ describes a new finding $F_{a}\not\in\{F_{i}\}$. Similarly, we simulate condition (b) through an exchange of finding where one finding sentence $S_{r}$ is removed to be replaced by another finding sentence $S_{a}$ as $R_{e}=\{S_{i}\}-\{S_{r}\}\cup \{S_{a}\}$. Finally, we can simulate the replacement of positive with negative findings and vice versa to form a revised report $R_{r}=\{S_{i}\}-\{S_{p}\}\cup \{S_{p'}\}$ where $S_{p}$ is a sentence corresponding to a finding $F_{p}$ and $S_{p'}$ is a sentence corresponding to the finding $F_{p'}$ which is in opposite sense of the meaning. For example, a sentence "There is pneumothorax", could be replaced by  "There is no pneumothorax" to represent a reversal of polarity of the finding.  Figure~\ref{fig3} shows examples of each of the type of operations of add, exchange and reverse findings respectively.

\subsection{Detecting findings in sentences}
\label{findingdetector}

Since detecting findings is key to our approach,  our synthetic dataset generation focused on chest X-ray datasets, as finding detectors are well-developed for these datasets. Further, the majority of work on automated reporting has been done on chest X-rays and finding-labeled datasets are publicly available \cite{NIHChestXray2017,mimic-4,irvin2019chexpert}. However, most of the existing approaches summarize findings at the report level. To locate findings at the sentence level, we used NLP tools such as Spacy\cite{negspacy} to separate sentences. We then combined ChexPert\cite{irvin2019chexpert} labeler and NegSpacy\cite{negspacy} parser to extract positive and negative findings from sentences. Table~\ref{tab1} shows examples of findings detected in sentences. The detected findings were then validated against the ground truth labels provided at the report level in the datasets. All unique findings across reports were then aggregated into a pool $\{F_{pool}\}$ and all unique sentences in the original reports were aggregated and mapped to their findings (positive or negative) to create the pool of sentences $\{S_{pool}\}$. 


\begin{table*}
\caption{Illustration of extracting findings from reports. Negated findings are shown within square brackets.}\label{tab1}
\begin{tabular}{l|l}
\toprule
{\bf Sentences} & {\bf Detected findings}\\
\midrule
There is effusion and pneumothorax. 	& `effusion', `pneumothorax'\\
\midrule
No pneumothorax, pleural effusion, but there is  & `consolidation',\\
 lobar air space consolidation. 	& [`pneumothorax', `pleural effusion']\\
\midrule
No visible pneumothorax or large pleural effusion. 	 &[`pneumothorax', `pleural effusion']\\
\midrule
Specifically, no evidence of focal consolidation, &[`focal consolidation'],\\
pneumothorax, or pleural effusion. 	& [`pneumothorax', `pleural effusion']\\
\midrule
No definite focal alveolar consolidation, & [`alveolar consolidation'],\\
no pleural effusion demonstrated. 	& [`pleural effusion']\\
\bottomrule
\end{tabular}
\end{table*}

\begin{table*}
\caption{Details of the fake report dataset distribution. 2557 frontal views were retained for images. 64 negative findings were retained and 114 positive findings.}\label{tab2}
\begin{tabular}{l|l|l|l|l|l|l|l}
\toprule
{\bf Dataset} & {\bf Patients} & {\bf Images} & {\bf Reports} & {\bf Pos/Neg}  & {\bf Unique}&{\bf Image-Sent} & {\bf Fake}\\
& & {\bf /Views}& & {\bf Findings}& {\bf Sentences} & {\bf Pairs} & {\bf Reports}\\
\midrule
Original & 1786 & 7470/2557 & 2557 & 119/64 & 3850 & 25535 & 7671\\ 
Training & 1071 & 2037 & 2037 & 68 & 2661 & 20326 & 4074 \\
Testing & 357 & 254 & 254 & 68 & 919 & 2550 & 508\\
\bottomrule
\end{tabular}
\end{table*}

\subsection{Fake report creation}
\label{fakereport}

For each original report $R$ associated with an image $I$, we create three instances of fake reports $R_{a}$, $R_{e}$, and $R_{r}$ corresponding to the operations of addition, exchange and reversal of findings, respectively. Specifically, for creating $R_{a}$ type of reports, we randomly draw from $S_{pool}$ a sentence that contains a randomly selected finding $F_{a}\notin\{F_{i}\}$, where $\{F_{i}\}$ are the set of findings in $R$ (positive or negative). Similarly, to create $R_{e}$, we randomly select a finding pair $(F_{ei},F_{eo})$, where $F_{ei}\in\{F_{i}\}$ and $F_{eo}\in\{F_{pool}\}-\{F_{i}\}$. We then remove the associated sentence with $F_{ei}$ in $R$ and replace it with a randomly chosen sentence associated with $F_{eo}$ in $\{S_{pool}\}$. Finally, to create the reversed findings reports, $R_{r}$, we randomly select a positive or negative finding $F_{p}\in\{F_{i}\}$, remove its corresponding sentence and swap it with a randomly chosen sentence $S_{p'}\in\{ S_{pool}\}$, containing findings $F_{p'}$ that is reversed in polarity. The images, their perturbed finding, and associated sentences were recorded in each case of fake reports so that they could be used to form the pairing dataset for training the fact-checking examiner described next. 

\section{Fact-checking of AI-generated reports}
\label{examine}

\subsection{Fact-checking Examiner}
The fact-checking examiner is a classifier using deep-learned features derived from joint image-text encodings. Specifically, since we combine images with textual sentences, we chose a feature encoding that is already trained on joint image and text pairs. In particular, we chose the CLIP joint image-text embedding model\cite{Radford2021} to project the image and textual sentences into a common 512-length encoding space. While other joint image-text encoders could potentially work, we chose CLIP as our encoder because it was pre-trained on  natural image-text  pair and subsequently tuned on radiology report-image pairs \cite{Endo2021}. We then concatenate the image and textual embedding into a 1024-length feature vector to train a binary classifier. In our splits, the real/fake incidence distribution was relatively balanced (2:1) so that the accuracy could be used as a reliable measure of performance.  We experimented with several classifiers ranging from support vector machines (SVM) to neural net classifiers. As we observed similar performance, we retained a simple linear SVM as sufficient for the task.

\subsection{Improving the quality of reports through verification}
\label{qi}
We apply the fact-checking examiner to filter our incorrect/irrelevant sentences in automatically produced reports as shown in Figure~\ref{fig:overview} (colored in red). Specifically, given an automatically generated report for an image, we pair the image with each sentence of the report. We then use the same CLIP encoder used in training the examiner,  to encode each pair of image and sentence to form a concatenated feature vector. The examiner predicted fake sentences are then removed to produce the revised report. 

We develop a new measure to judge the improvement in the quality of the automatic report after applying the fact-checking examiner. Unlike popular report comparison measures such as BLEU\cite{papineni-etal-2002-bleu}, ROUGE\cite{ROUGE} scores that perform lexical comparisons, we use a semantic similarity measure formed from encoding the reports through large language models such as SentenceBERT\cite{sbert}.  Specifically, let $R=\{S_{i}\}$, $R_{auto}=\{S_{auto}\}$, $R_{corrected}=\{S_{corrected}\}$ be the  original, automated, and corrected reports with their sentences respectively.  To judge the improvement in quality of the report, we adopt SentenceBERT\cite{sbert} to encode the individual sentences of the respective reports to produce an average encoding per report as $E_{R},E_{auto},E_{corrected}$ respectively. Then the quality improvement score, $QI(R)$ per triple of reports $(R, R_{auto}, R_{corrected})$ is given by the difference in the cosine similarity between the pairwise encodings  as
\begin{equation}
QI(R,R_{auto},R_{corrected})=d(E_{R},E_{corrected})-d(E_{R},R_{auto}),
\end{equation} 
\noindent where $d$ is the cosine similarity between the average encodings. This measure allows for unequal lengths of reports. A positive value indicates an improvement while a negative value indicates a worsening of the performance. 
The overall improvement in the quality of automatically generated reports is then given by
\begin{equation}
QI=(n_{positive}+n_{same}-n_{negative}) / n_{R}
\end{equation}
\noindent where
\begin{equation}
\begin{split}
 n_{positive}=|{arg_{R}(d(E_{R},E_{corrected})>d(E_{R},R_{auto}))}|\\
 n_{same}=|{arg_{R}(d(E_{R},E_{corrected})=d(E_{R},R_{auto}))}|\\ 
 n_{negative}=|{arg_{R}(d(E_{R},E_{corrected})<d(E_{R},R_{auto}))}|
 \end{split}
\end{equation}
are the number of times the corrected reports are closer to original reports, same similarity as the AI report, or worse than the AI report by applying the examiner respectively, and $n_{R}$ is the total number of automated reports evaluated. 

\section{Results}
\label{results}

To test our approach for fact-checking of radiology reports,  we selected an open access dataset of  chest X-rays from Indiana University\cite{indiana}  provided on Kaggle, which contains 7,470 chest X-Ray (frontal and lateral views) images with corresponding 2,557 non-duplicate reports from 1786 patients. The dataset also came with annotations  documenting important findings at the report level. Of the 1786 patients, we used a (60-20-20)\% patient split for training the examiner, testing the examiner, and evaluating its effectiveness in report correction respectively, thus ensuring no patient overlap between the partitions.


\begin{table}[t]
\begin{minipage}{0.48\textwidth}
\begin{centering}
\caption{Performance of real/fake discriminator.}\label{tab:performance}
\begin{tabular}{c|l|l|l|l}
\toprule
& {\bf Prec.} & {\bf Recall} & {\bf F1} & {\bf Sup.}\\
\midrule
Real class & 0.86 & 0.93 & 0.90 & 3648 \\
Accuracy & -- & -- & 0.84 & 5044 \\
Macro Avg & 0.82 & 0.77 & 0.79 & 5044\\
Weighted Avg & 0.84 & 0.84 & 0.84 & 5044\\
\bottomrule
\end{tabular}
\end{centering}
\end{minipage}
\begin{minipage}{0.52\textwidth}
\begin{centering}
    \includegraphics[width=.9\textwidth]{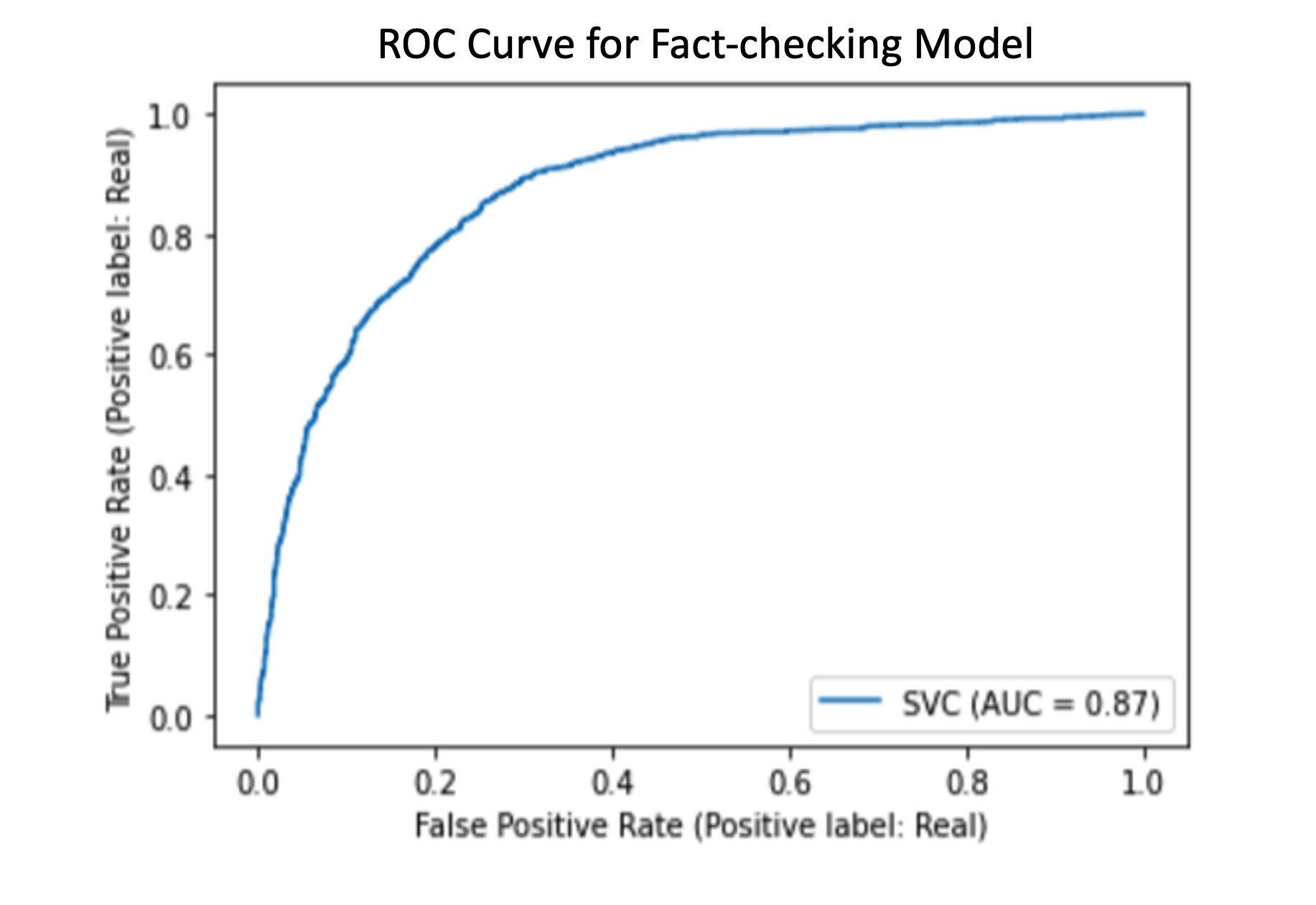}
\captionof{figure}{performance of real/fake report sentence differentiation.}
\label{fig5}
\end{centering}
\end{minipage}
\end{table}
\subsection{Fake report dataset created}
By applying NLP methods of sentence extraction, we extracted 3850 unique sentences from radiology reports. By applying the finding extractor at the sentence level as described in Section~\ref{findingdetector}, we catalogued a total of 119 distinct positive and 64 negative findings as shown Table~\ref{tab2}. Using these findings and their sentences in the 2557 unique reports, and the 3 types of single perturbation operations described in Section~\ref{findingtypes}, we generated 7,671 fake reports as shown in Table~\ref{tab2}.

The training and test dataset for the fact-checking examiner was generated by randomly drawing sentences from sentence pool $\{S_{pool}\}$.  Each image was first paired with each sentence from its original report and the pair was given the "Real" label. The perturbed sentence  drawn from $\{S_{pool}\}$ from the fake reports was then retrieved from each fake report and paired with the image and given the "Fake" label. The list of pairs produced were processed to remove duplicate pairings.   By this process, we generated 20,326 pairs of images with real/fake sentences for training, and 2,550 pairs for testing as shown in Table~\ref{tab2} using 80\% of the 1,786 patients. 


\subsection{Fact-checking examiner accuracy}
Using the train-test splits shown in Table~\ref{tab2}, we trained fact-checking examiner with encodings of image-sentence pairs shown in Table~\ref{tab2}. The resulting classifier achieved an average accuracy of 84.2\% and the AUC was 0.87 as shown in Figure~\ref{fig5}b. The precision, recall, F-score, macro accuracy by average and weighted average methods are shown in Figure~\ref{fig5}a.  By using 10 fold cross-validation in the generation of the (60-20-20) splits for the image-report dataset, and using different classifiers provided in the Sklearn library (decision tree, logistic regression, etc.) the average accuracy lay in the range $0.84 \pm 0.02$. The errors seen in the classification were primarily for complex description of findings containing modifiers and negations. 

\subsection{Overall report quality improvement evaluation}
We evaluated the efficacy of the fact-checking examiner on two report datasets, one synthetic with controlled ``fakeness" and another dataset generated by a published algorithm described in \cite{syeda-mahmood2020}. Specifically, using the 20\% partition of patients from the Indiana reports that was not used to train or test the examiner, we selected 3089 of the fake reports shown in Table~\ref{tab2}. We evaluated the improvement in report quality using the method described in Section~\ref{qi}.  These results are summarized in Table~\ref{tab:evaluation}. Since our fake reports had only one fake sentence added, the performance improvement while still present, is modest around 15.63\%.

\begin{table}[t]
\centering
\caption{Report quality evaluation on two automatically generated report datasets.}\label{tab:evaluation}
\begin{tabular}{l|c|c|c|c|c|c}
\toprule
{\bf Dataset} &{\bf Patients}  & {\bf Reports} & {\bf $n_{positive}$} & {\bf $n_{same}$} &{\bf $n_{negative}$}& {\bf QI score} \\
\midrule
Synthetic Reports & 358 & 3661 & 1105 & 1008 & 1548 & 15.63\%\\
\midrule
NIH Reports & 198 & 198 & 60 & 55 & 83 & 16.1\%\\
\bottomrule
\end{tabular}
\end{table}

To test the performance on automated reports generated by existing algorithms, we obtained a reference dataset consisting of freshly created reports on the NIH image dataset\cite{NIHChestXray2017} created by radiologists as described in \cite{Wu2020}. We retained the output of an automated report generation algorithm for the same images described in \cite{syeda-mahmood2020}.
A total of 198 pairs of original and automatically created reports along with their associated imaging from the NIH dataset was used for this experiment. The results of quality improvement is shown in Table~\ref{tab:evaluation}. As it can be seen, the quality improvement is slightly greater for reports produced by automated report extraction methods. 

\section{Conclusion}

In this paper, we have proposed for the first time, an image-driven verification of automatically produced radiology reports. A dataset was carefully constructed to elicit the different types of errors produced by such methods. A Novel fact-checking examiner was developed using pairs of real and fake sentences with their corresponding imaging. The work will be extended in future to cover larger variety of defects and extended evaluation on a larger number automated reports. 
%
%
\bibliography{anthology, refs_py}
\bibliographystyle{splncs04}
\end{document}